\documentclass[10pt,twocolumn,letterpaper]{article}

\usepackage{iccp}
\usepackage{times}
\usepackage{amsmath}
\usepackage{amssymb}
\usepackage{url}
\usepackage{graphicx}
\graphicspath{{figures/}}
\usepackage{tikz}
\usepackage{tikzscale}
\usepackage{units}
\usepackage{subfigure}

\usepackage{chngcntr}
\usepackage{stmaryrd}
\usepackage{gensymb} 
\usepackage[capitalize,noabbrev]{cleveref}
\crefformat{equation}{(#2#1#3)}
\crefname{section}{\S}{\S}
\usepackage{etoolbox}
\usepackage[multiple]{footmisc}

\renewcommand{\vec}[1]{\ensuremath{\mathbf{#1}}}
\DeclareMathOperator{\argmin}{argmin}
\newcommand{\dd}[1]{\ \mathrm{d}{#1}}
\usepackage{booktabs}
\usepackage{multirow}

\iccpfinalcopy 
\newtoggle{cvww}
\togglefalse{cvww}
\def\iccpPaperID{3} 

\ificcpfinal\fi

\begin{document}
\title{Real-Time Panoramic Tracking for Event Cameras}

\author{Christian Reinbacher, Gottfried Munda and Thomas Pock
\\
Institute for Computer Graphics and Vision, Graz University of Technology
\\
{\tt\small \{reinbacher,gottfried.munda,pock\}@icg.tugraz.at}
}

\maketitle
\iftoggle{cvww}
{\ifcvwwfinal\thispagestyle{fancy}\fi}
{\thispagestyle{empty}}

\begin{abstract}
  Event cameras are a paradigm shift in camera technology.
  Instead of full frames, the sensor captures a sparse set of {\em events} caused by intensity changes.
  Since only the changes are transferred, those cameras are able to capture quick movements of
  objects in the scene or of the camera itself.
  In this work we propose a novel method to perform camera tracking of event cameras in a panoramic setting with three degrees of freedom.
  We propose a direct camera tracking formulation, similar to state-of-the-art in visual odometry.
  We show that the minimal information needed for simultaneous tracking and mapping is the spatial position of events, without using the appearance of the imaged scene point.
  We verify the robustness to fast camera movements and dynamic objects in the scene on a recently proposed dataset~\cite{Mueggler2016} and self-recorded sequences.
\end{abstract}

\section{Introduction}
Event cameras such as the Dynamic Vision Sensor (DVS)~\cite{Lichtsteiner2008} work asynchronously on a pixel level which is in stark contrast to standard CMOS digital cameras that operate on frame basis.
Each pixel measures the incoming light intensity individually and fires an {\em event} when the absolute change in intensity is above a certain threshold.
The time resolution is in the order of $\mu s$ and the bandwidth of commercially available cameras is up to \unit[1.000.000]{events/s}.
Due to the sparse nature of the events, the amount of data that has to be transferred from the camera to the computer is very low, making it an energy efficient alternative to standard CMOS cameras for the tracking of very quick movements \cite{Delbruck2007,Wiesmann2012}.
Due to the asynchronous nature of events, computer vision algorithms that operate on a frame basis cannot be readily applied.
Very recently, the computer vision community started to investigate the applicability of those cameras to classical computer vision problems (\eg camera pose estimation ~\cite{Benosman2014,Gallego2015,Kim2014,Kim2016,Mueggler2015,Weikersdorfer2013}), leveraging the high temporal resolution and low latency of event cameras.

\begin{figure}
  \begin{center}
    \includegraphics[width=0.3\columnwidth]{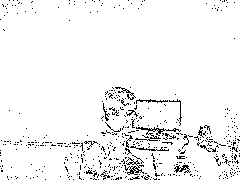}
    \includegraphics[width=0.6\columnwidth]{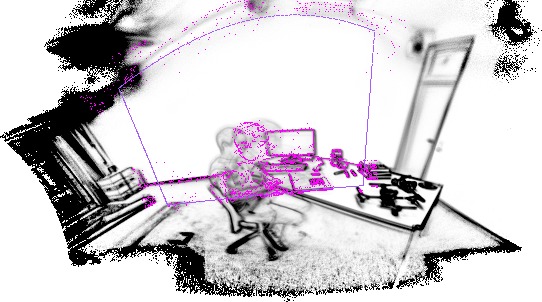}\\
    \iftoggle{cvww}{
      \subfigure[Raw Events]{\includegraphics[width=0.3\columnwidth]{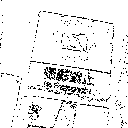}\label{subfig:events}}
      \subfigure[Reconstructed Map]{\includegraphics[width=0.6\columnwidth]{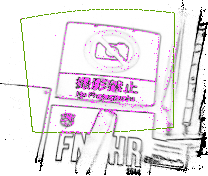}\label{subfig:image}}
    }{
      \subfigure[Raw Events]{\includegraphics[width=0.3\columnwidth]{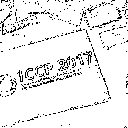}\label{subfig:events}}
      \subfigure[Reconstructed Map]{\includegraphics[width=0.6\columnwidth]{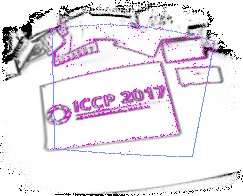}\label{subfig:image}}
    }
  	\caption{Sample results from our method.
    The left column shows the raw events (images only created for visualization) and right column the resulting map created by our tracking method.
    The current camera pose and input events are visualized in color. Note, that the person in the first image is moving.}
  	\label{fig:teaser}
  \end{center}
  \vspace*{-12pt}
\end{figure}

\paragraph*{Contribution}
In this work we propose a panoramic mapping and tracking method that only relies on the geometric properties of the event stream, the event position in the image.
We show how ideas from dense camera tracking can be applied to this new domain, leading to a simultaneous tracking and mapping method that can be easily implemented to run in real-time with several hundred pose updates per second.
Both tracking and mapping are defined on a per-event-basis.
The design of the mapping function naturally lends itself to super-resolution (essential for the resolution of currently available cameras) and can also handle dynamic objects in the scene (as can be seen in \cref{fig:teaser}).
We show the tracking accuracy of our method on a recently proposed dataset \cite{Mueggler2016} which provides event-streams with temporally aligned ground-truth camera orientation measurements.
We have implemented our method in an interactive application for which we provide source code.

\section{Related Work}
Event-based cameras receive increasing interest from robotics and computer vision researchers.
The high temporal resolution make them interesting for tracking rapid movements of either objects or the camera itself.
Also low-level computer vision problems are transferred to this new domain like optical flow estimation \cite{Benosman2014,Bardow2016}.
In this overview we focus on recent work that aims to solve camera pose tracking and mapping using this new camera paradigm.
Simultaneous localization and mapping (SLAM) methods that are based on visual input need to perform image  matching to build a map of the environment and localize the camera within (see \eg~\cite{Hartmann2013} for a survey).
Since event cameras do not readily provide an image, the vast majority of visual SLAM algorithms can not be applied directly.
Milford \etal~\cite{Milford2015} were among the first to show that feature extraction from artificial images, created by accumulating events over time slices of \unit[1000]{ms} (similar to \cref{subfig:events}), allows to perform large-scale mapping and localization with loop-closure.
While this is the first system to utilize event cameras for this challenging task, it is inspired by classical frame-based computer vision approaches, thus trading temporal resolution for the creation of images to reliably track camera movement.

Ideally, camera pose updates should be formulated on an event basis.
Cook \etal~\cite{Cook2011} propose a biologically inspired network that simultaneously estimates camera rotation, image gradients and intensity information.
Similarly, Kim \etal~\cite{Kim2014} propose a method to simultaneously estimate the camera rotation around a fixed point and a high-quality intensity image only from the event stream.
A particle filter is used to integrate the events and allow a reconstruction of the image gradients, which are then used to reconstruct an intensity image by Poisson editing.
Both approaches require the reconstructed intensity image to formulate the camera pose update.
In contrast to that, an indoor application of a robot navigating in 2D using an event camera that observes the ceiling has been proposed by Weikersdorfer \etal~\cite{Weikersdorfer2013}.
They simultaneously estimate a 2D map of events and track the 2D position and orientation of the robot by just utilizing the spatial position of the events.
All methods are limited to 3 DOF of camera movement and utilize a filtering approach to update the camera pose.
A full camera tracking has been shown in \cite{Mueggler2014,Mueggler2015} for rapid movement of an UAV with respect to a known 2D target and in \cite{Gallego2015} for a known 3D map of the environment (including information about the intensity for each 3D point).
Kim \etal~\cite{Kim2016} recently proposed a full SLAM pipeline for event cameras, featuring 6 DOF tracking and map creation by augmenting their previous method \cite{Kim2014} with a pixel-wise inverse depth filter working on a key-frame basis.

In contrast to the presented methods, in this work we will take a step back and identify the smallest amount of information necessary to perform camera tracking from a sparse event stream.
In the following section we will present
a direct method that optimized re-projection errors between a known map of events and the current event.
We show this on the application of panoramic camera tracking with three rotational degrees of freedom.
The same principles can be used to extended the method to full 6 DOF\footnote{During the review phase of this paper, this has already been shown by Rebecq \etal~\cite{Rebecq2016a}. They combine a recently proposed method for depth estimation for event cameras \cite{Rebecq2016} with camera pose estimation based on similar techniques as in the presented work.}.

\section{Panoramic Event Camera Tracking}
Our method for panoramic camera tracking is inspired by direct alignment approaches like DTAM~\cite{Newcombe2011}, LSD-SLAM~\cite{Engel2014} and DSO~\cite{Engel2016}.
Those methods pose the camera pose estimation problem as minimization of a photometric error between the current camera frame and a known map of the world as
\begin{equation}\label{eqn:semidense_errfun}
  \min_\theta{\frac{1}{2}\sum_{\vec{x}\in\Omega}\delta(\vec{x})\left(I(\vec{x})-M(\phi(\vec{x},\theta)) \right)^2}\ ,
\end{equation}
where $\Omega\subset\mathbb{R}^2$, $\theta$ being some parameterization of the current camera position in the world, and  $\phi(\vec{x},\theta)$ being a projection function that maps an image point $\vec{x}\in\Omega$ to a point in a map $M$.
$\delta(\vec{x}):\Omega \to \{0,1\}$ is an indicator function that selects a subset of pixels to be included in the error function.
DTAM~\cite{Newcombe2011} uses all pixels in the current frame for alignment, while LSD-SLAM and DSO~\cite{Engel2014,Engel2016} operate on a subset of pixels around large image gradients (potential edges and corners).
While in aforementioned approaches $\delta(\vec{x})$ has to be defined heuristically, the sparse output of event cameras corresponds to intensity changes caused by color differences in the scene.

More formally, the output of an event camera is a time sequence of events $(e^n)_{n=1}^N$, where $e^n=\{x^n,y^n,p^n,t^n\}$ is a single event consisting of the pixel coordinates $(x^n,y^n)\in \Omega$, the polarity $p^n\in \{ -1,1\}$ and a monotonically increasing timestamp $t^n$.
An event will be fired once the change in log-intensity space has crossed a certain threshold $\pm C$.
Methods which use an intensity image for tracking (\eg~\cite{Gallego2015,Kim2014,Kim2016}) optimize for apparent camera motion under brightness constancy assumption where the event formation is given by
\begin{equation}\label{eqn:intensity_tracking}
  |\langle \nabla \log I,\vec{u}\Delta t \rangle| \geq C\ ,
\end{equation}
with $\vec{u}$ being a 2D motion field and $\Delta t$ the time difference between two events at the same image position.
This requires the knowledge of $\Delta t$ and $C$ which in practice are subject to noise of the imaging sensor.
To overcome wrong predictions of \cref{eqn:intensity_tracking}, \cite{Kim2014,Kim2016} embed it into a {\em Particle Filter} framework while \cite{Gallego2015} use an {\em Extended Kalman Filter}.
In this work we drop the information contained in $p^n$ and utilize only the position $(x^n,y^n)$ of an event $e^n$ for camera tracking.

In the following sections we define the mapping function and how to generate the map given known camera positions in \cref{sec:mapping}.
In \cref{sec:tracking} we show how the camera pose can be estimated given a map $M$.
Finally we propose a practical method to simultaneous camera tracking and mapping in \cref{sec:slam}.
\subsection{Mapping}\label{sec:mapping}
In this section we describe the mapping process in our algorithm.
We consider a two-dimensional map $M:\Gamma\to [0,1],\Gamma\subset\mathbb{R}^2$ and we will give statistical justification of the quantity that will be stored in our map.
For each position $v\in\Gamma$, the map should represent a likelihood of an event occurring at that position when the event camera is moved over it.
Let us assume that we have given a panoramic image $I$ of a static scene and we have given a 1D path $\mathcal{S}\subset\Gamma$, describing the motion of pixels induced by the camera movement over the map.
Furthermore we shall assume that the path $\mathcal{S}$ is decomposed into a finite number of segments $s_i\in\mathbb{R}^2,i=1\dots n$, each of length $r$.

According to \cref{eqn:intensity_tracking} we can define the number of events that will be triggered at each point $v\in\Gamma$ as
\begin{align}\label{eqn:map_counting}
  \#e(v|\mathcal{S}) = \sum_{i=1}^n & \left\llbracket|\langle \nabla \log I(v),s_i\rangle|>C\right\rrbracket \nonumber \\
  & \llbracket(v)\in W(s_i)\rrbracket\ ,
\end{align}
where $W(s_i)$ refers to a suitable windowing function centered around the segment $s_i$, indicating if a map pixel $v$ is currently visible by the camera and $\llbracket \dotsc \rrbracket$ are indicator functions.

Without loss of generality we assume that $W(s_i)=const$\footnote{For example by a small random motion of the camera around a certain point of the map.}, such that for all $v\in W$ we have that
\begin{equation}\label{eqn:map_counting_visible_pixels}
  \frac{\#e(v|\mathcal{S})}{n} = \frac{1}{n}\sum_{i=1}^n\left\llbracket|\langle \nabla \log I(v),s_i\rangle|>C\right\rrbracket\ ,
\end{equation}
where we have divided both sides of the equation by $n$. Observe that by passing $n$ to infinity, the left hand side will converge to the probability $P(e|v,\mathcal{S})$ of a pixel $v$ to trigger an event $e$.

\begin{figure}
  \centering
  \newlength{\svgwidth}
  \setlength{\svgwidth}{0.7\columnwidth}%
  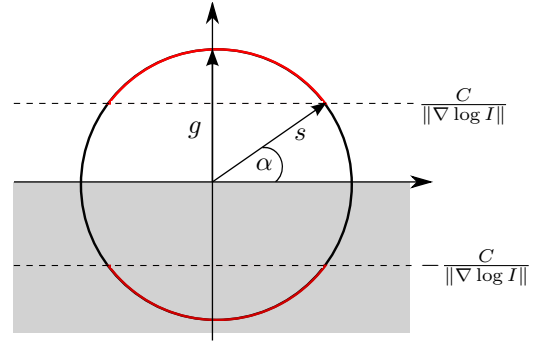
  \caption{Graphical derivation of the probability of an event given threshold $C$ for a horizontal edge in image $I$. $g=\frac{\nabla \log I}{\|\nabla \log I\|}$ is the unit gradient vector. $P(e|v)$ corresponds to the ratio of arc length, depicted in red, to the full circle.}
  \label{fig:dot_product}
\end{figure}

Our goal is to gain more insight into the relation between the probability $P(e|v)$ (\ie the probability with the camera motion $\mathcal{S}$ factored out) and the panoramic image $I$.
To this end we assume that the camera motion is represented by a random variable $s$, uniformly sampled on the unit circle $\mathcal{S}^1$, that is $P(s)=1/(2\pi)$.
Hence the event probability can be written as
\begin{equation}
  P(e|v)=\int_{S^1}\left\llbracket|\langle \nabla \log I(v),s\rangle|>C\right\rrbracket P(s) \dd s,
\end{equation}
Observe that the critical angle $\alpha$ between $\nabla \log I^\perp$ and the motion $s$ for which $\langle\nabla \log I,s\rangle=C$ is given by
\begin{equation}
  \alpha = \sin^{-1}\left(\frac{C}{\|\nabla \log I\|}\right)\ .
\end{equation}
With that we can calculate the probability for an event as the arc length of the circle for which the condition $\llbracket|\langle \nabla \log I,s\rangle|>C\rrbracket$ holds as
\begin{align}\label{eqn:event_prob}
  P(e|v) &= (2 \pi - 4\alpha) P(s) \nonumber \\
       &= 1 - \frac{2}{\pi} \sin^{-1}\left(\frac{C}{\|\nabla \log I(v)\|}\right)\ ,
\end{align}
which is depicted in \cref{fig:dot_product}.
This formula shows that the event-probability in a certain pixel $v$ is proportional to the gradient magnitude of the {\em log}-intensity image.
Finally, we also point out that since $I>0,\|\nabla \log I\|=\|\nabla I\|/I$.
Hence $P(e|v)$ is proportional to Weber's law of physical stimuli~\cite{Shen2006}.

Next, we show how we can approximate $P(e|v)$ by only considering event positions, which is in contrast to some recently proposed work such as \cite{Kim2014,Kim2016}.
We derived that the number of events that have been triggered at a certain map position \cref{eqn:map_counting} resembles an event probability which is proportional to the gradient magnitude of the {\em log}-intensity image \cref{eqn:event_prob} when the camera movement is factored out.
Thus our goal is now to calculate \cref{eqn:map_counting} for a single segment of the camera path.
We therefore define our map using a method proposed by Weikersdorfer \etal~\cite{Weikersdorfer2013}, originally intended for 2D SLAM using a robotic platform with upward-facing camera.
Formally, we represent the map as a normalized version of \cref{eqn:map_counting}
\begin{equation}
  M(v)=\frac{O(v)}{N(v)}\ ,
\end{equation}
where $O:\Gamma\to\mathbb{N}^0$ counts the number of event occurrences at position $v\in\Gamma$ and $N:\Gamma\to\mathbb{R}^+$ the number of possible occurrences.
The update of $O$ for an incoming event at camera position $\vec{x}$ can be straightforwardly defined as
\begin{equation}\label{eqn:update_occurences}
  O(\phi(\vec{x},\theta_t)) = O(\phi(\vec{x},\theta_t)) +1\ ,
\end{equation}
where $\phi(\vec{x},\theta_t)$ is the projection of event position $\vec{x}$ into the map using the current camera estimate $\theta_t$.
In contrast to that, \cite{Weikersdorfer2013} define a probabilistic update which utilizes the current belief of a particle filter.

$N$ normalizes for the camera path segment, ending at map pixel $v$.
In practice the segments $s_i$ defined in \cref{eqn:map_counting} are not of equal length. Therefore we define this term as the length of the camera path segment that ended at $v$.
We can compute this as
\begin{equation}\label{eqn:update_normalization}
  \underset{\vec{x}\in\Omega}{N(\phi(\vec{x},\theta_t))} = N(\phi(\vec{x},\theta_t)) + \|\phi(\vec{x},\theta_t)-\phi(\vec{x},\theta_{t-1})\|\ .
\end{equation}
Note that the normalization map has to be updated for all camera pixels, \ie all map pixels $\in W(s_i)$, while the occurrence map is only updated for the current event.
Two examples of resulting maps $M$ are depicted in \cref{subfig:image}.
To complete the description of the map, we define the projection function as
\begin{equation}\label{eqn:projection}
  \phi(\vec{x},\theta) = \begin{bmatrix}\renewcommand*{\arraystretch}{1.8}
    p_x(1+\frac{\tan^{-1}(\nicefrac{X_x}{X_z})}{\pi})\\
    p_y(1+\frac{X_y}{\sqrt{X_z^2+X_x^2}})
  \end{bmatrix}\ ,
\end{equation}
where $(X_x,X_y,X_z)^T$ are the components of $\vec{X} = R_\theta K^{-1}(x,y,1)^T$, $K$ is the camera calibration matrix, $R_\theta$ is the rotation matrix generated by an axis-angle representation $\theta$ and $(p_x,p_y)^T$ is the center point of $M$.
\cref{eqn:projection} projects a camera pixel $\vec{x}$ for camera rotation $\theta$ to a cylindrical coordinate system, especially well suited for $360\degree$ panoramas since there is not much distortion in $x$-direction~\cite{Szeliski1997}.
\subsection{Tracking}\label{sec:tracking}
Given a map as defined in \cref{sec:mapping}, we seek to iteratively update the current camera pose for new events.
For that, we define the re-projection error of a single event akin to \cref{eqn:semidense_errfun} as
\begin{equation}\label{eqn:single_event_error}
  g(\theta,e^n)=1-M(\phi(\vec{x},\theta))\ ,
\end{equation}
with $\vec{x}=(x^n,y^n)$ and $M$ being a panoramic map of event probabilities in the environment, defined in \cref{sec:mapping}.
We pose the camera tracking as the energy minimization problem
\begin{equation}\label{eqn:tracking_objective}
    \theta_t = \argmin_\theta{\frac{1}{2}\left\|g(\theta,\vec{e}^n) \right\|^2}\ + \frac{\alpha}{2}\|\theta - \theta_{t-1}\|^{2},
\end{equation}
A single event does not contain enough information to allow for optimization, so we always process packets of events $\vec{e}^n$. The second term ensures that the estimated camera movement stays small.
To optimize \cref{eqn:tracking_objective}, we employ a Gauss-Newton-type scheme and
replace the residual $r(\theta)=g(\theta,\vec{e}^n)$ by a linearized version

\begin{equation} \label{eqn:linearized_objective}
  r(\theta)\approx r(\theta_0)+J(\theta-\theta_0)\ ,
\end{equation}
with $\theta_0 = \theta_{t-1}$ and $J$ being the Jacobian of the projection function.
\iftoggle{cvww}{}{A detailed derivation of $J$ can be found in \cref{sec:suppl_detail}.}
We see that inserting \cref{eqn:linearized_objective} into \cref{eqn:tracking_objective}
gives a linear least-squares problem with the solution
\begin{align}\label{eqn:update_standard}
  \theta_{k+1} &= \theta_{k} -(J_k^T J_k + \alpha I)^{-1} (J_k^T r(\theta_k)\nonumber \\
  &-\alpha(\theta_{k}-\theta_{t-1}))\ ,
\end{align}
where we use $k$ to denote the inner iterations in contrast to the outer problem denoted by $t$.
To further increase robustness and speed of convergence, we perform the following update steps known
as Nesterov accelerated gradient descent \cite{Nesterov1983}
\begin{subequations}\label{eqn:update_accelerated}
  \begin{align}
    \vartheta_k &= \theta_k + \beta (\theta_k-\theta_{k-1})\\
    \theta_{k+1} &= \vartheta_k - (J_k^T J_k+\alpha I)^{-1} (J_k^T r(\vartheta_k)\nonumber \\
    &-\alpha(\vartheta_{k}-\theta_{t-1}))\ ,
  \end{align}
\end{subequations}
where $\beta = [0,1]$ and $J_k$ is evaluated at $\vartheta_k$.
In practice, less than $k=10$ iterations of \cref{eqn:update_accelerated} are sufficient to reach convergence.
\subsection{Simultaneous Tracking and Mapping}\label{sec:slam}
In the previous sections we introduced our proposed methods for camera tracking and map creation.
For a practical system it is essential that the map creation and camera tracking can be run simultaneously.
 We combine both modules as described as follows.
 The incoming event stream is divided according to the time stamps into packets of $\delta t$.
 For each packet a pose update $\theta_t$ is computed according to \cref{sec:tracking}.
 If the update is successful, the new pose is in turn used to update the map according to \cref{eqn:update_occurences,eqn:update_normalization} with all events in the packet.
 We update the map after every single pose update with all available events.
 In cases where tracking becomes unreliable (high residual of \cref{eqn:tracking_objective}), we switch off the map update in order not to corrupt the map with incorrect updates.

\paragraph*{Implementation Details}
For bootstrapping the system, we simply switch off the pose updates for the first few packets, such that the map gets populated.
The map is updated quickly, so errors made in the initialization do not have a large influence on the tracking result.
We implement map update and gradient calculation on GPUs, since both can be easily parallelized and are carried out in every iteration.
The optimization is running on the CPU.

\section{Experiments}
For our experiments we utilize two different event cameras, the DVS128~\cite{Lichtsteiner2008} with a spatial resolution of $128\times 128$ and a DAVIS240~\cite{Brandli2014} with a resolution of $240\times 180$. Since we do not have access to the latter camera, we use a recently proposed data-set by Mueggler \etal~\cite{Mueggler2016} which was recorded using a DAVIS240 camera.

\subsection{Calibration}
Our camera tracking model requires an intrinsically calibrated camera. The DVS can only sense intensity differences, which are either caused by moving objects, a moving camera in a static scene or a change in illumination. The first two options are not feasible for camera calibration. Instead, we replace the usual printed calibration target by an LCD which displays a picture of the calibration target, quickly alternating with a blank frame. This results in images that can be used as an input to traditional camera calibration pipelines. Alternatively, one could also use an image-reconstruction method for event cameras (\eg~\cite{Bardow2016,Reinbacher2016}). We use the automatic feature detection of the calibration framework of Ferstl \etal~\cite{Ferstl2015}, specifically designed for low-resolution cameras, together with a MATLAB camera calibration toolbox\footnote{\url{http://www.vision.caltech.edu/bouguetj/calib_doc}}. We release the source for an automatic calibration tool suited for event cameras along with this publication\footnote{\url{https://github.com/VLOGroup/dvs-calibration}}.

\subsection{Tracking Accurary}
\begin{figure}
  \centering
  \subfigure[Poster]{\includegraphics[width=0.48\columnwidth]{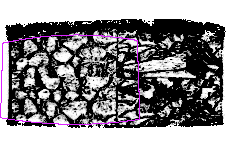}}
  \subfigure[Boxes]{\includegraphics[width=0.48\columnwidth]{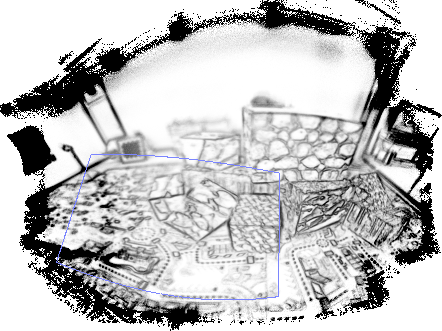}}\\
  \subfigure[Shapes]{\includegraphics[width=0.48\columnwidth]{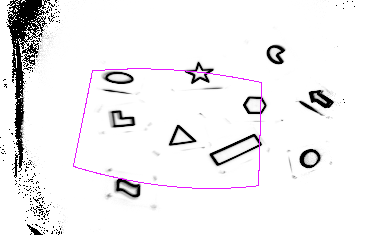}}
  \subfigure[Dynamic]{\includegraphics[width=0.48\columnwidth]{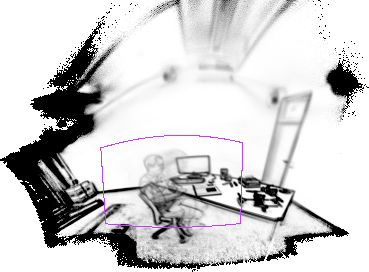}}
  \caption{Sample output of our algorithm on the tested datasets. The current camera pose is visualized as a colored rectangle.}
  \label{fig:dataset_qualitative}
\end{figure}

In this section we show the robustness of the proposed tracking method on a variety of sequences that were recorded using a DAVIS240 camera~\cite{Mueggler2016}.
\paragraph*{Dataset}
The dataset contains 27 sequences recorded in 11 scenes. Out of those sequences, four contain only rotational camera movement and are therefore usable to assess the quality of the proposed panoramic camera tracking method.
The sequences used are \unit[60]{s} long and contain between \unit[17]{million} and \unit[180]{million} events.
Ground-truth camera poses are available with a time resolution of \unit[200]{Hz}.
The pose measurements are synchronized to the event stream as well as the recorded DAVIS frames (which are not used in this work).
The four sequences ({\em Poster}, {\em Boxes}, {\em Shapes} and {\em Dynamic}) are depicted in \cref{fig:dataset_qualitative} by means of reconstruction output of our method.
All sequences start with relatively slow rotational motion which intensifies up to \unit[880]{\degree} per second for the {\em Poster} sequence. {\em Poster} and {\em Boxes} are recorded in a heavily textured environment, whereas {\em Shapes} and {\em Dynamic} provoke comparatively few events.

\paragraph*{Evaluation Protocol}
The authors of \cite{Mueggler2016} do not provide an evaluation protocol for their dataset.
We therefore define the tracking error to be the angle between ground-truth and estimated camera viewing direction.
While this error metric neglects in-plane-rotation it can nevertheless be used to compare different parametrizations of our method.
The ground truth consists of 6 DOF camera positions in the coordinate system of the external tracking device used.
To compare our estimate, we ignore the translation that is present in the ground truth since it is small in comparison to the distance between camera and scene objects.
Furthermore, we temporally align our output to the ground truth and estimate a global rotation between the two coordinate systems using RANSAC.
We note that during the review phase of this paper, Gallego and Scaramuzza \cite{Gallego2016a} have proposed a method that estimates angular velocities of moving event cameras.
Their method replaces the global map by contrast maximization of images generated from a short time frame of events.
Therefore their focus is on angular velocity estimation rather than global position estimation.
While the method is evaluated on the same dataset, we refrain from a direct comparison due to this different focus.

\begin{figure}[t]
  \centering
  \subfigure[Shapes]{\includegraphics[width=0.48\columnwidth]{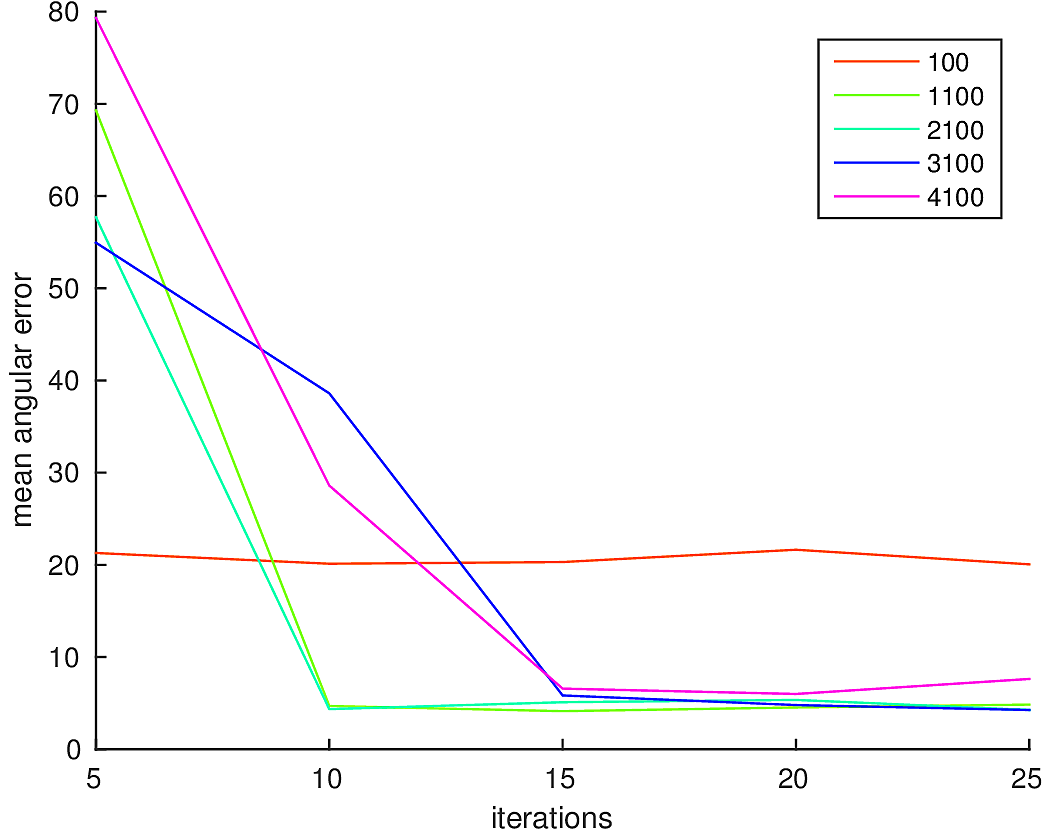}}
  \subfigure[Poster]{\includegraphics[width=0.48\columnwidth]{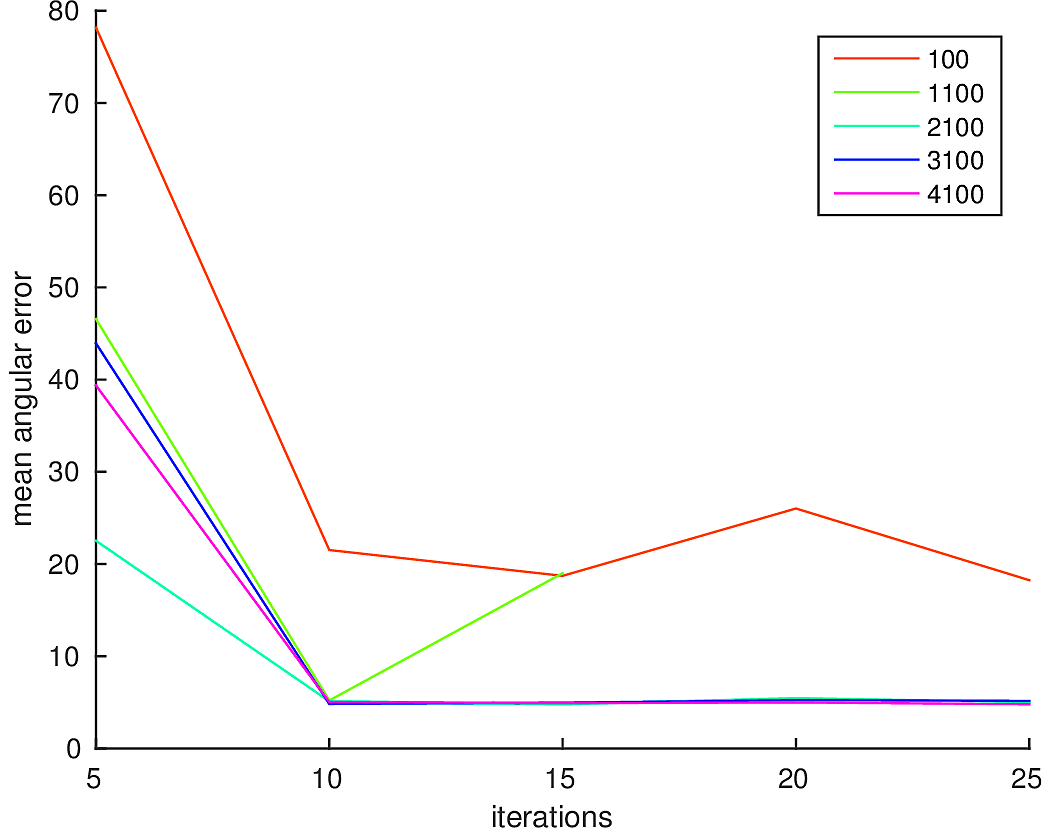}}
  \caption{Influence of the number of iterations and event packet size on the tracking accuracy of our method. We compare a highly textured scene (Poster) with a rather low texture scene (Shapes). }
  \label{fig:num_events}
\end{figure}

\paragraph*{Influence of Hyperparameters}
The hyperparameters of our method are number of events per packet and the number of iterations per packet. We keep the remaining parameters $\alpha=1$ and $\beta=0.4$ fixed throughout all experiments.
We start our investigation by relating the number of events in one packet with the number of iterations necessary.
In \cref{fig:num_events} we plot the mean angular error (as defined above) over the number of iterations in the optimization for different packet sizes.
While we only depict two datasets in \cref{fig:num_events}, the results on all four datasets indicate that $10$ optimization iterations per packet are sufficient to achieve a good tracking performance, with a mean angular error below $5^\circ$.
Iterating further does not significantly decrease the error, indicating that the optimization already converges for $10$ iterations.
The event packet size is related to the amount of texture in the scene.
For highly textured scenes, using larger packet sizes might be beneficial to achieve real-time performance (see \cref{sec:timing} for a detailed timing analysis).
Our method is rather agnostic to the packet size regarding the tracking accuracy.
A packet size of $1000$ to $2000$ events (corresponding to \unit[$\approx$2-5]{ms}) is a good compromise between time resolution and tracking performance.

\begin{figure}[t]
  \centering
  \includegraphics[width=0.8\columnwidth]{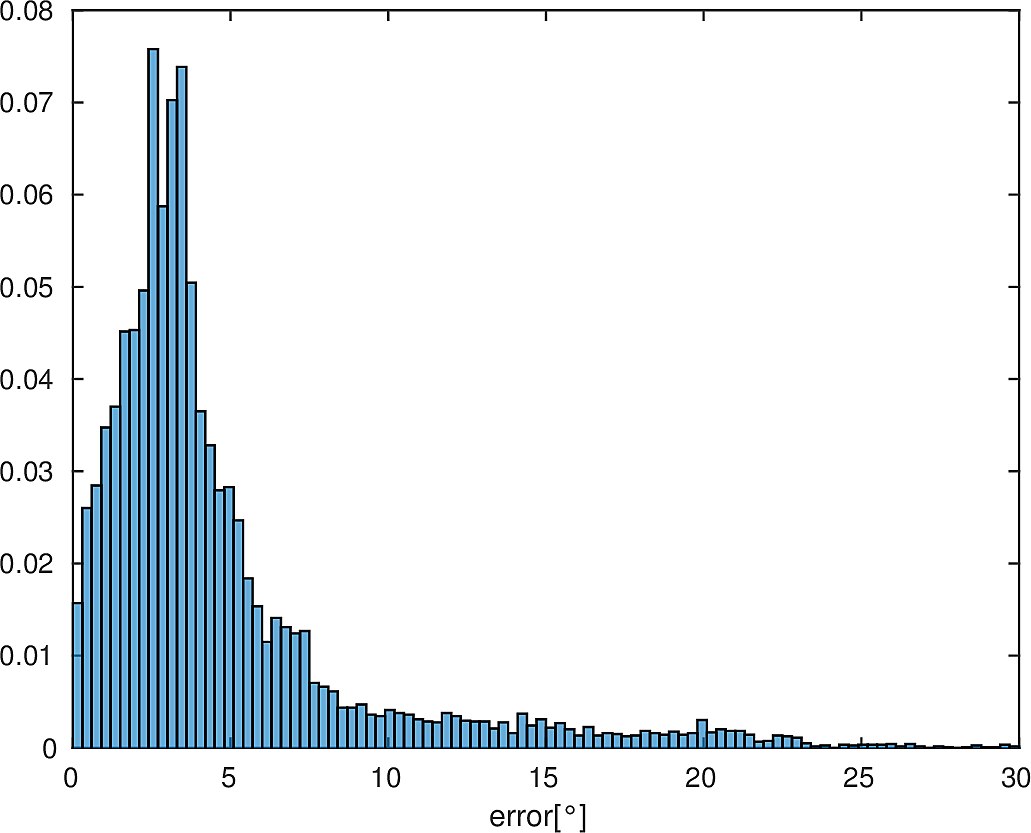}
  \caption{Histogram of the tracking error on the {\em Poster} sequence. The majority of frames can be tracked with an error below $5^\circ$.}
  \label{fig:poster_gt}
\end{figure}

To give more insight into the behavior of the method, we additionally visualize the histogram of the tracking error for the {\em Poster} sequence in \cref{fig:poster_gt}.
As can be seen from this quantitative evaluation, the proposed method is able to accurately track the camera orientation for slow as well as fast camera shake.
Our method sometimes underestimates the motion for very fast camera motion, however it can quickly catch up and successfully continue tracking.
Note, that our method has no explicit relocalization, but thanks to the global map of the scene is able to recover from temporary tracking failures.

\begin{figure}[t]
  \includegraphics[width=\columnwidth]{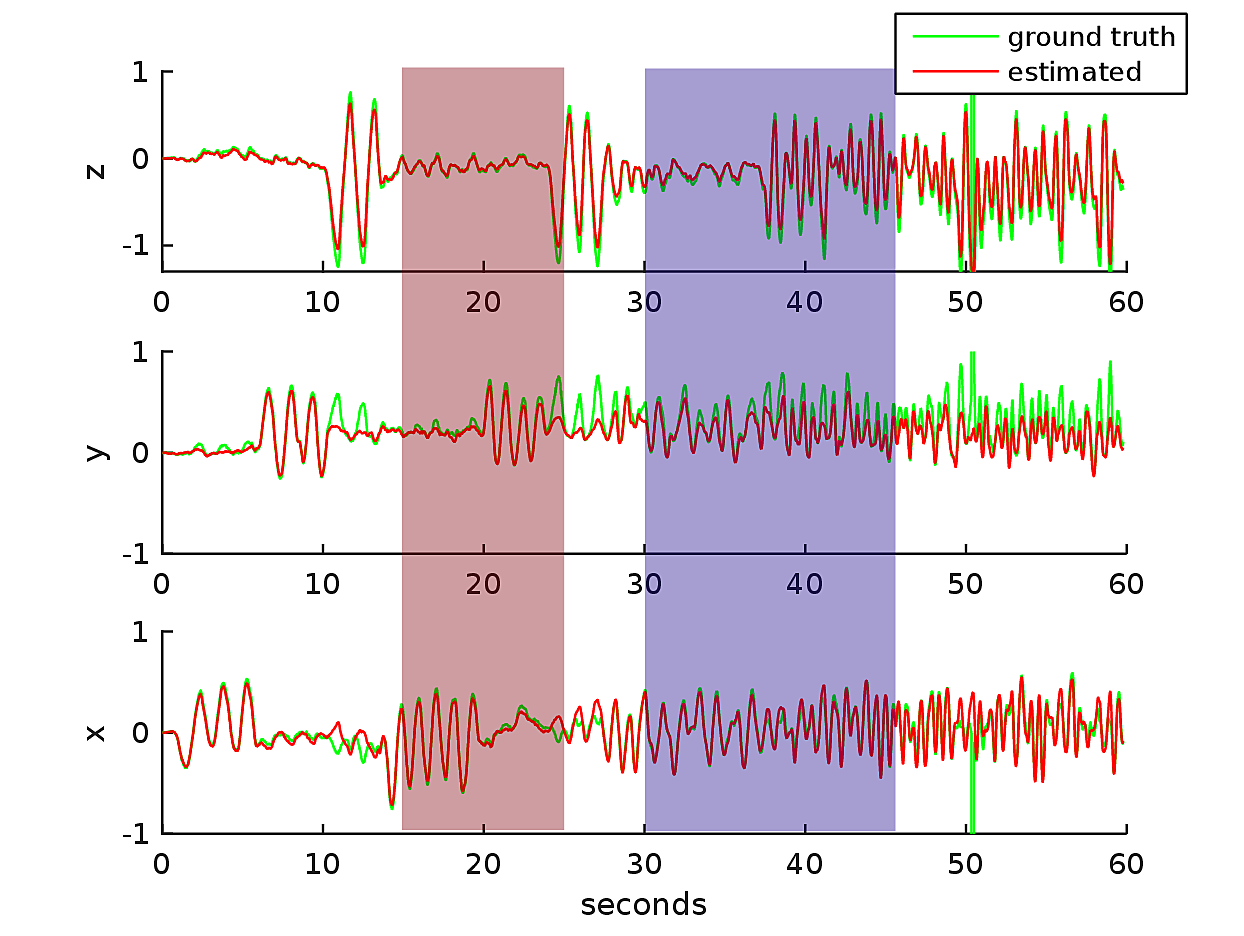}
  \caption{Comparison to ground truth on the {\em Dynamic} sequence over the whole duration of \unit[60]{s}. Green curve depicts the recorded ground truth of an external tracking system, red is the output of our method. During the red phase, the person starts moving, during the blue phase the person stands up and walks around the table.}
  \label{fig:dynamic}
\end{figure}

\paragraph*{Dynamic Scenes}
The {\em Dynamic} sequence is especially challenging since the depicted person starts to move after a few seconds.
This violates our model which assumes the scene to be static.
We analyze the impact in \cref{fig:dynamic} where we plot the angular tracking error over time.
During the phase marked in red, the person starts moving, in the blue phase the person stands up.
An example of this movement is depicted in \cref{fig:teaser}.
Due to that movement the tracking quality decreases. Nevertheless our method is able to successfully track through the complete sequence\footnote{\url{https://youtu.be/Qy0brSlirmk} shows the same visualization related to the live input.}.

\begin{figure}
  \centering
  \subfigure[1.0]{\includegraphics[width=0.48\columnwidth]{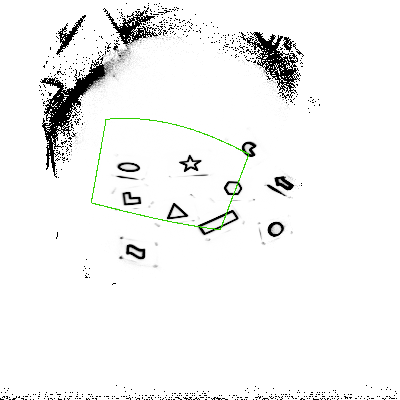}}
  \subfigure[1.2]{\includegraphics[width=0.48\columnwidth]{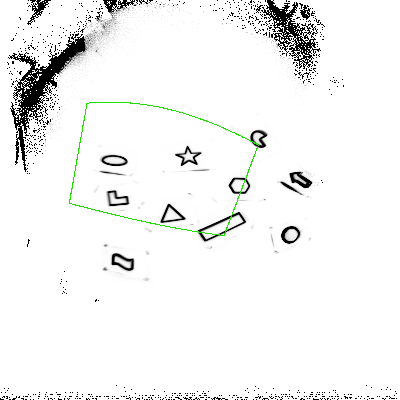}}\\
  \subfigure[1.5]{\includegraphics[width=0.48\columnwidth]{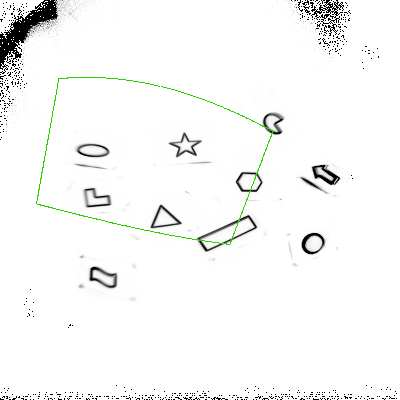}}
  \subfigure[2.0]{\includegraphics[width=0.48\columnwidth]{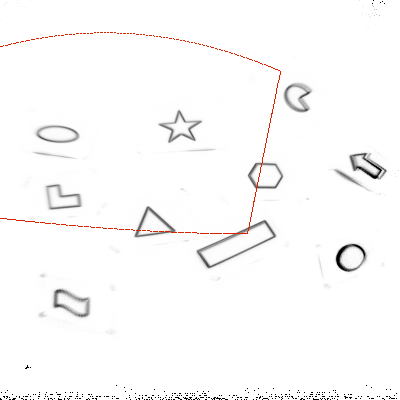}}
  \caption{Super resolution on the {\em Shapes} dataset. The numbers correspond to the upsampling factor. Tracking and reconstruction work reliably up to an upsampling factor of $1.5$.}
  \label{fig:superresolution}
\end{figure}

\paragraph*{Super Resolution}
The resolution of the panorama $M$ is completely independent from the resolution of the camera. Pose tracking naturally works with sub-pixel resolution which allows us to create a panorama with a much higher resolution. In \cref{fig:superresolution} we show outputs created with upsampling factors of $1$, $1.2$, $1.5$ and $2$. As we can see, the quality of the reconstruction is worse for large upsampling factors, however a factor of up to $1.5$ is feasible in practice. To use higher upsampling factors, it is neccessary to add a blur operator to the map update in \cref{eqn:update_occurences}.

\begin{figure*}[t!]
  \centering
  \includegraphics[width=\textwidth]{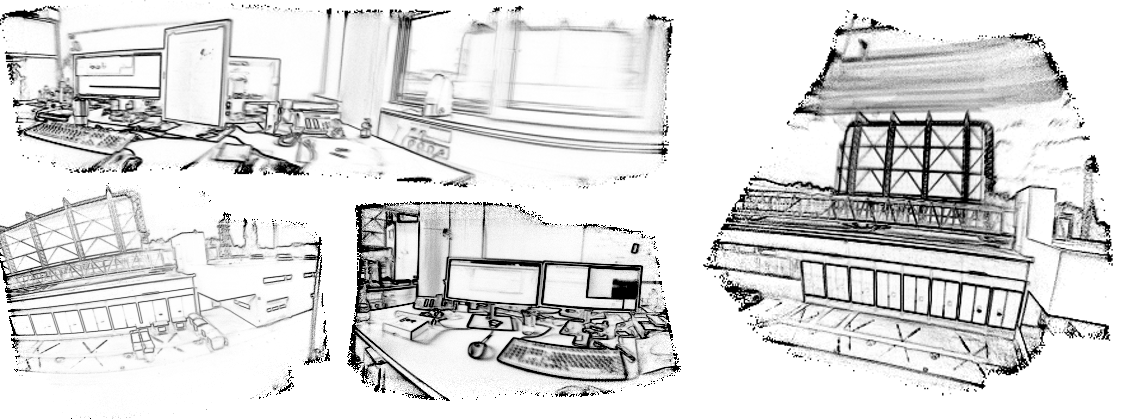}
  \caption{Panoramic maps produced by our method, recorded using a DVS128 camera.}
  \label{fig:qualitative}
\end{figure*}
\subsection{Timing}\label{sec:timing}
In this section we investigate the influence of parameter choices on the runtime of our algorithm.
For these experiments we utilize data from a DVS128.
The low resolution of the camera makes it especially difficult to be used for tracking.
We have captured several sequences around our office, inside and outside.
The resulting reconstructions are depicted in \cref{fig:qualitative}.

All timings have been recorded using a Linux PC with a 3.3 GHz Intel Core i7 and a NVidia GTX 780 ti GPU.
In \cref{tbl:runtime} we report the runtime of the individual components mapping and tracking for different event package sizes.
For the optimization, only the time for a single iteration is reported, since it scales linearly with the number of iterations.
The results in \cref{fig:qualitative} are created using a packet size of $1500$ and $10$ iterations, which allows for reliable tracking of rapid camera movement. This leads to $\approx 170$ pose updates per second.
\begin{table}
  \centering
  \caption{Runtime analysis of the individual components of our method in \unit{ms}.}
  \vspace*{4pt}
  \label{tbl:runtime}
  \begin{tabular}{lccc}
    \toprule
    \textbf{\#Events} & \textbf{Mapping} & \textbf{Tracking}\\
    \textbf{ /packet} & [ms] & [ms/iter]\\
    \midrule
    500 & 0.3 & 0.38 \\
    1000 & 0.31 & 0.49 \\
    1500 & 0.48 & 0.56 \\
    2000 & 0.5 & 1.01 \\
    2500 & 0.51 & 1.05 \\
    \bottomrule
  \end{tabular}

\end{table}
\section{Conclusion}
In this work we proposed a novel method for performing panoramic mapping and tracking using event cameras.
In contrast to related work which uses particle filtering for camera tracking \cite{Kim2014,Weikersdorfer2013}, we formulate the tracking as an optimization problem that can be efficiently solved at real-time frame rates on modern PCs.
The presented mapping method accurately represents for each position the probability of events being generated, independent from the direction of the current camera movement.
In extensive experiments we have shown the robustness of our tracking method to rapid camera movement and dynamic objects in the scene.
The source code for the developed real-time framework is available online\footnote{\url{https://github.com/VLOGroup/dvs-panotracking}}.


\subsection*{Acknowledgements}
  This work was supported by the research initiative Intelligent Vision Austria with funding
  from the AIT and the Austrian Federal Ministry of Science, Research and Economy
  HRSM programme (BGBl. II Nr. 292/2012).

{\small
\bibliographystyle{ieee}
\bibliography{egbib}
}
\iftoggle{cvww}{}
{
\clearpage
\newpage
\appendix

\setcounter{figure}{0}
\setcounter{table}{0}
\counterwithin{figure}{section}
\counterwithin{table}{section}

\twocolumn[{%
 \centering
 \LARGE Real-Time Panoramic Tracking for Event Cameras \\ (Appendix) \\[1.5em]
 \normalsize
}]

\section{Technical Details}\label{sec:suppl_detail}
\subsection{Jacobian of the Projection Function}
In \cref{sec:tracking} we optimize for the camera position by minimizing a linearized version of \cref{eqn:tracking_objective}. We start the derivation of the full Jacobian by rewriting the projection function \cref{eqn:projection} as
\begin{equation}
  \phi(\vec{x},\theta) = \mathcal{P}(g(G(\theta),\bf{X}))\ ,
\end{equation}
where $\vec{X}=K^{-1}\vec{x}$ and $G(\theta) = e^{[\theta]_\times}$ is the exponential map that takes the elements of the
Lie-algebra to the corresponding manifold. In the following, $[\cdot]_\times$ denotes the operator
that generates a skew-symmetric matrix from a vector:
\begin{equation}
  \label{eq:1}
  \begin{bmatrix}
x_1\\x_2\\x_3
\end{bmatrix}_\times =   \begin{bmatrix}
  0 & -x_3 & x_2 \\
  x_3 & 0 & -x_1 \\
  -x_2 & x_1 & 0
  \end{bmatrix}
\end{equation}
  $g(T,\vec{X}) = T\vec{X}$ transforms the 3D-point $\vec{X}$ using the matrix $T\in SE(3)$ and $\mathcal{P}$ is the spherical projection function defined in \cref{eqn:projection}.
For the optimization of \cref{eqn:linearized_objective} we are interested in the derivative
\begin{align}\label{eqn:jacobian}
  &\left. \frac{\partial M(\phi(\vec{x},\theta))}{\partial \theta}\right|_{\theta=\theta_0} = J = \nonumber \\
  &\left. \frac{\partial M}{\partial \vec{\tilde x}}\right|_{\vec{\tilde x}=\phi(\vec{x},\theta_0)}
  \left. \frac{\partial \mathcal{P}(\vec{\tilde X})}{\partial \vec{\tilde X}}\right|_{\vec{\tilde X} = g(G(\theta_0),\vec{X})}\nonumber\\
  &\left. \frac{\partial g(\tilde G(\vec{X}))}{\partial \tilde G}\right|_{\tilde G = G(\theta_0)}
  \left. \frac{\partial G(\theta)}{\partial \theta}\right|_{\theta=\theta_0}\ .
\end{align}
Let us now define each therm in \cref{eqn:jacobian}. The derivative of the map $M$ will simply be denoted as $\nabla M$, where $\nabla$ is the gradient operator in the image domain $\Omega$. The partial derivate of the camera rotation is given by
\begin{equation}
  \left. \frac{\partial G(\theta)}{\partial \theta}\right|_{\theta=\theta_0} =
  \begin{bmatrix}
    -[r_1]_\times\\
    -[r_2]_\times\\
    -[r_3]_\times
  \end{bmatrix},
\end{equation}
where $r_i$ denotes the $i$-th column of the rotation matrix $R=e^{[\theta_0]_\times}$.
The derivation of the transformation is defined as
\begin{equation}
  \left. \frac{\partial g(\tilde G(\vec{X}))}{\partial \tilde G}\right|_{\tilde G = G(\theta_0)} =
  \vec{X}^T \otimes I_{3\times 3},
\end{equation}
with $I_{3 \times 3}$ the $3\times 3$ identity matrix and $\otimes$ the Kronecker matrix product.

Finally, the derivative of the projection function is
\begin{align}
  \left. \frac{\partial \mathcal{P}(\vec{\tilde X})}{\partial \vec{\tilde X}}\right|_{\vec{\tilde X} = g(G(\theta_0),\vec{X})} =\nonumber \\
  \renewcommand*{\arraystretch}{1.5}
  \begin{bmatrix}
    -\frac{p_x X_x}{\pi (X_z^2+X_x^2)} & \frac{p_x X_z}{\pi (X_z^2+X_x^2)} & 0\\
    -\frac{p_y X_z X_y}{(X_z^2+X_x^2)^{\nicefrac{3}{2}}} &
    -\frac{p_y X_x X_y}{(X_z^2+X_x^2)^{\nicefrac{3}{2}}} & -\frac{p_y}{\sqrt{X_z^2+X_x^2}}
  \end{bmatrix}\ ,
\end{align}
which concludes the derivation of the full Jacobian. We do not have to store the full $J$ matrix
since we are only interested in $J^{T}J$ which can be computed incrementally by summing up the outer
product of the rows $J_j$ of $J$ (one residuum corresponds to one row of $J$)
\begin{equation}
  J^TJ = \sum_{j=1}^N J_{j}^{T}J_{j}
\end{equation}


}

\end{document}